\documentclass[10pt,twocolumn,letterpaper]{article}

\usepackage{wacv}
\usepackage{times}
\usepackage{epsfig}
\usepackage{graphicx}
\usepackage{amsmath}
\usepackage{amssymb}
\usepackage{subcaption}
\usepackage{multirow}

\wacvfinalcopy 

\def\wacvPaperID{****} 

\ifwacvfinal\fi
\begin{document}

\title{Style and Content Disentanglement in Generative Adversarial Networks}

\author{Hadi Kazemi \\
West Virginia University\\
{\tt\small hakazemi@mix.wvu.edu}
\and
Seyed Mehdi Iranmanesh \\
West Virginia University\\
{\tt\small seiranmanesh@mix.wvu.edu}
\and
Nasser M. Nasrabadi \\
West Virginia University\\
{\tt\small nasser.nasrabadi@mail.wvu.edu}
}

\maketitle
\ifwacvfinal\thispagestyle{empty}\fi

\begin{abstract}
Disentangling factors of variation within data has become a very challenging problem for image generation tasks. Current frameworks for training a Generative Adversarial Network (GAN), learn to disentangle the representations of the data in an unsupervised fashion and capture the most significant factors of the data variations. However, these approaches ignore the principle of content and style disentanglement in image generation, which means their learned latent code may alter the content and style of the generated images at the same time. This paper describes the Style and Content Disentangled GAN (SC-GAN), a new unsupervised algorithm for training GANs that learns disentangled style and content representations of the data. We assume that the representation of an image can be decomposed into a content code that represents the geometrical information of the data, and a style code that captures textural properties.
Consequently, by fixing the style portion of the latent representation, we can generate diverse images in a particular style. Reversely, we can set the content code and generate a specific scene in a variety of styles. The proposed SC-GAN has two components: a content code which is the input to the generator, and a style code which modifies the scene style through modification of the Adaptive Instance Normalization (AdaIN) layers' parameters. We evaluate the proposed SC-GAN framework on a set of baseline datasets. 
\end{abstract}

\section{Introduction}

Learning visual representations from large unlabeled data has been an area of active research in computer vision. In this context, the goal is to learn a representation that describes the remarkable semantic features of an image. A method that can learn such representation may be adopted by a variety of supervised learning tasks such as visualization, regression, and classification. Generative models, and more specifically Generative Adversarial Networks (GANs), with no doubt, are among the most powerful techniques of unsupervised representation learning. The underlying belief of the generative frameworks is that the ability to synthesize an observed data encompasses some sort of understanding.

In practice, however, GANs are not able to learn a meaningful representation of the training dataset without additional constraints. To sidestep this problem, a long line of work proposed different frameworks to learn interpretable and meaningful latent representations in an unsupervised setting, such as InfoGAN~\cite{chen2016infogan}, BiGAN~\cite{donahue2016adversarial}, and ALI~\cite{dumoulin2016adversarially}, or a supervised setting \cite{8296650, perarnau2016invertible}. 

Despite all the effort in this area, these approaches ignore one of the most fundamental principles of image generation, which is the disentanglement of the scene's content and style. A scene's content, here, represents its underlying geometry, while the style encodes the texture and illumination. Wang et al. \cite{wang2016generative} propose to decomposes the GAN latent code into two separate structure and style codes. However, their method requires the depth information and does not generalize to RGB datasets. In contrast, in this paper we propose a universal GAN framework, which does not need the depth information, to learn disentangled content and style codes. Factoring the style and content enhance the interpretability of the learned representations, compared to the counterparts, improves the performances of learning tasks which are reliant on the sole content or style representation, and enables the user to generate a specific scene with a variety of styles, or several images of a particular style. 

Learning disentangled style and content codes is a challenging task. Several algorithms have been developed for solving this class of problems, however, they train GAN in a supervised fashion, or in its conditional setting. In addition, these frameworks address the problem of image-to-image translation, that is an image of a source domain, e.g., edges, represents the content, and a style code is learned for the target domain. Despite all the effort in this area, the field still lacks a coherent framework for unsupervised disentangled style and content representations learning. From this consideration, in this paper, we propose an end-to-end framework to learn a pair of disentangled content and style codes for a given dataset. Note that our framework makes the assumption that each image of the training dataset can be decomposed into content and style codes.

Our framework consists of a single generator, in contrast to \cite{wang2016generative}, whose input is the content code. The generator has several upsampling, convolutional layers, and a set of residual blocks. Inspired by a recent work which showed that parameters of affine transformation in normalization layers represent the styles, we equip each residual block with an Adaptive Instance Normalization (AdaIN) layer. The parameters of the AdaIN layers then generated by a multilayer perceptron (MLP) whose input is the desired style code. We develop a training strategy and discuss the required considerations for its success, to train the MLP and generator jointly. Note that, with minimum effort, the proposed framework can be adopted by any GAN structure. Finally, by learning to invert the generator, similar to BiGAN \cite{donahue2016adversarial}, we can scale up the applications of our framework to transferring style or content among images of the training domain. 

The main contributions of our work are fourfold:
\begin{itemize}
	\item We present the Style and Content Disentangled GAN (SC-GAN) that learns to disentangle style and content representations of the data in an unsupervised fashion.
	\item We propose a training scheme based on the proposed losses to train the proposed SC-GAN.
	\item We extend the BiGAN model using the proposed SC-GAN framework, which enables us to transfer style and content between data samples.
	\item We present qualitative, quantitative experiments showing that the proposed framework can effectively disentangle the style and code and improve the performance of supervised tasks.
\end{itemize}

\section{Related Work}
\subsection{Generative Adversarial Network (GAN)}
Building generative models, that are able to model high-dimensional data distributions, is a fundamental problem within many computer vision applications, such as face generation \cite{karras2017progressive}, image-to-image translation \cite{pix2pix2017}, image editing \cite{ding2017exprgan}, image in-painting \cite{yeh2017semantic}, and speech synthesis \cite{Hao2018CMCGANAU}. Currently, the most prominent approaches are generative adversarial networks (GAN) \cite{goodfellow2014generative}, Variational Autoencoders (VAE) \cite{kingma2013auto}, and Auto-Regressive Generative Models \cite{salimans2017pixelcnn++}.  These models capture the joint distribution between the data and a set of hidden variables, called latent codes, representing different variations of the data. The trained models then generate new samples, in the training domain, given random latent codes, which are sampled from their prior distributions. Prior works conditioned these models on additional information to direct the data generation process. The conditioning could be on another image for image-to-image translation, part-of-image for inpainting, some desired data attributes \cite{yan2016attribute2image}, or even class labels \cite{mirza2014conditional}. Although these methods produce impressive photorealistic images, they fail to learn an interpretable representation of the data.

InfoGAN \cite{chen2016infogan}, an information-theoretic extension of GAN, allows learning of representation which is partially interpretable. The resulting code then consists of a meaningful part corresponding to specific semantic attributes of the data, and a random part which injects diversity among the generated samples. In contrast, two concurrent independent works \cite{donahue2016adversarial, dumoulin2016adversarially} proposed a full inference of the random code. They have demonstrated that these codes can learn the semantic attributes of the data. Several other papers have also investigated supervised representation learning by conditioning the discriminator on specific attributes \cite{8296650,perarnau2016invertible}. Also, transferring attributes among images has been studied in the literature \cite{ding2017exprgan,huang2018multimodal}.

Despite their success, they overlook the key principle of image generation, which is the style and content disentanglement. Learning a distinct style and content code is studied in \cite{zhu2017toward, lee2018diverse} for image-to-image translation. However, they do not learn the underlying content code and use an image of another domain as the reference content. Wang et al. \cite{wang2016generative} study this problem in unconditional setting to learn a disentangle representation of the data. However, their method utilizes the depth information to learn the content, which they refer to as structure representation.

\subsection{Neural Style Transfer}
Recently several methods were developed to transfer style between images. Style transfer is a technique which enables rendering texture from a reference image while preserving the semantic content of the target image. The early work by Gatus et al. \cite{gatys2016image} showed that a deep
neural network (DNN) can encode both the style and content information and proposed an iterative method to transfer the style of an artistic image to an arbitrary photograph. To accelerate neural style transfer, some techniques are proposed to perform stylization using feed-forward neural networks in a single forward pass \cite{johnson2016perceptual, ulyanov2016texture, li2016precomputed}. However, these techniques are restricted to a single style and cannot adapt to an unseen arbitrary style. To sidestep this issue, Dumoulin et al. \cite{dumoulin2017learned} propose conditional instance normalization to learn the normalization parameters of distinct styles. Li et al. \cite{li2017diversified} also guide the network to synthesize the desired style using a texture selector network. However, all these methods are either limited to transferring a few sets of styles, or too slow for real-time applications. 

The Instance Normalization (IN) \cite{ulyanov2017improved} has been found to carry the style information of an image \cite{li2017demystifying, gatys2016image}. Inspired by the IN success, Haung et al. \cite{huang2017arbitrary} proposed the Adaptive Instance Normalization (AdaIN) to adjust the mean and variance of the content features with those of the style features.  Given feature activations of the content and style images, Chen et al. \cite{chen2016fast} replaced content features with the closest-matching style features patch-by-patch. A universal stylization is also presented in \cite{li2017universal} based on the whitening and coloring transform which stylizes images via feature projection.

\section{Preliminaries}
In this section, we provide some rudiments of GANs and style transfer, necessary to understand the proposed SC-GAN framework.
\subsection{Generative Adversarial Networks (GANs)}\label{sec:GAN}
GANs \cite{goodfellow2014generative} are a group of generative models which learn the statistical distribution of training data, allowing us to synthesize data samples by mapping a random noise $z$ to an output image $y$: $G(z): z \longrightarrow y$, where $G$ is the generator network. GAN in its conditional setting (cGAN) is proposed in \cite{isola2016image} which learns a mapping from an input $x$ and a random noise $z$ to the output image $y$: $G(x, z): \{x, z\} \longrightarrow y$, using an autoencoder network (generator). The generator model, $G(x, z)$, is trained to generate an image which is not distinguishable from "real" samples by a discriminator network, $D$. Simultaneously, the discriminator is learning, adversarially, to discriminate between the "fake" generated images by the generator and the real samples from the training dataset. Consequently, the GAN objective function is defined as 
\begin{align} \label{eq:ad_loss}
l_{d}(G,D) &=  \mathbf{E}_{x,y\sim p_{data}}[\log D(x,y)] \\ \nonumber &+  \mathbf{E}_{x, z\sim p_{z}}[\log (1 - D(x,G(x, z)))],
\end{align}
where $G$ attempts to minimize it and $D$ tries to maximize it.

\subsection{Style Transfer}\label{sec:style}
Replacing Batch Normalization (BN) layers with IN layers can significantly improve the style transfer networks \cite{ulyanov2017improved}. Given $x\in R^{N\times C \times H \times W}$, an input tensor containing a batch of $N$ images of size $H \times W$ with $C$ channels, IN is given by: 
\begin{align} \label{eq:IN}
IN(x) &= \gamma \Big ( \dfrac{x-\mu(x)}{\sigma(x)}\Big ) + \beta ,
\end{align}
where $\gamma, \beta$ are affine parameters learned from data, $\mu(x)$ and $\sigma(x)$ are computed across spatial dimensions independently for each channel and each sample:
\begin{align} \label{eq:mu}
\mu_{nc}(x) &= \dfrac{1}{HW} \sum_{h=1}^{H} \sum_{w=1}^{W} x_{nchw},
\end{align}
\begin{align} \label{eq:sigma}
\sigma_{nc}(x) &= \sqrt{\dfrac{1}{HW} \sum_{h=1}^{H} \sum_{w=1}^{W}. (x_{nchw}-\mu_{nc}(x))^2 + \epsilon }.
\end{align}
Note that, in contrast to BN, which usually replaces mini-batch statistics with population statistics at the test time, IN remains similar to its training time. Instance normalization performs a form of style normalization by normalizing feature statistics, namely the mean and variance. In other words, it normalizes the input to a single style specified by its affine parameters. To adapt it to arbitrarily given styles, Haung et al. \cite{huang2017arbitrary} proposed Adaptive IN (AdaIN) which employs adaptive affine transformations:
\begin{align} \label{eq:AdaIN}
AdaIN(x, y) &= \sigma(y) \Big ( \dfrac{x-\mu(x)}{\sigma(x)}\Big ) + \mu(y) ,
\end{align}
where $x$ is a content input and $y$ is a style input. Unlike IN, AdaIN has no affine parameter to learn. It simply scale and shift the normalized content input by $\sigma(y)$ and $\mu(x)$, respectively.

\section{Proposed Style and Content Disentangled GAN (SC-GAN)}
\subsection{Network Architecture} \label{sec:network}
\begin{figure*}[t]
	\begin{center}
		
		\includegraphics[width=0.85\linewidth]{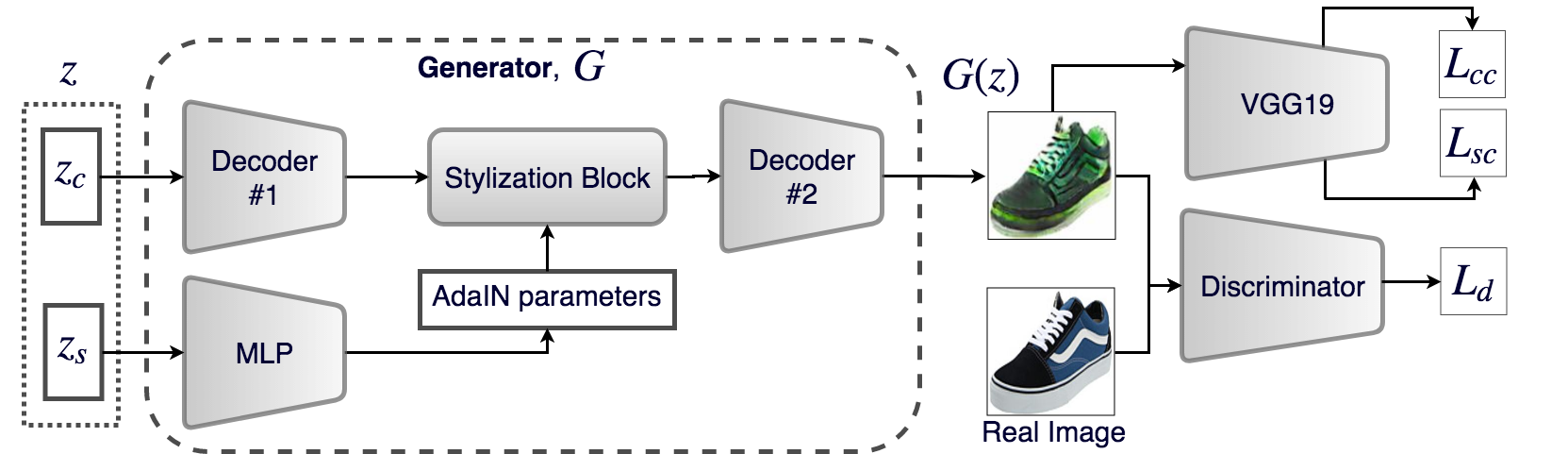}
	\end{center}
	\caption{The proposed SC-GAN framework, consists of a generator $G$, a discriminator $D$, and a pre-trained VGG19 to extract features for content and style consistency losses. The MLP dynamically changes the parameters of AdaIN layers in stylization block to render different styles on the synthesized image.}
	\label{fig:framework}
\end{figure*}

Our generator, $G$, consists of 4 different blocks: decoder \#1, residual blocks, decoder \#2 and an MLP (see Figure~\ref{fig:framework}). The first decoder processes the content code by several upsampling and convolutional layers. All the convolutional layers of this decoder are followed by IN. 
The stylization block comprises a set of residual blocks which are responsible for imposing the requested style on the output image. Inspired by recent works that use affine transformation parameters in normalization layers to represent styles \cite{huang2017arbitrary}, we equip the residual blocks with AdaIN layers whose parameters are dynamically generated by the MLP block from a style code.
Finally, the decoder \#2 construct an image by several upsampling and convolutional layers. Since IN removes the original feature mean and variance that represent essential style information, the second decoder is equipped with batch normalization. 

To adopt SC-GAN by any GAN framework, the generator is the only part which needs to be modified. In other words, the discriminator could be left intact. Consequently, we do not change the original discriminator of the GAN frameworks that we are using in this paper.

\subsection{Training the SC-GAN}
The proposed SC-GAN takes a random code $z = (z_c, z_s)$ composes of a content code $z_c$ and a style code $z_s$ as input, and synthesizes an output image, $G(z)$. 
However, the network needs a mechanism to learn relating the content of the generated image to the content code and its style to the style code. In particular, the content of the generated image is supposed to be intact as long as $z_c$ remains unaltered, despite the value of $z_s$, and vice versa. In order to train the generator to learn these relations, a pair of content codes, $z_{c1}, z_{c2}$, and a pair of style codes, $z_{s1}, z_{s2}$, are drawn randomly. Now, we generate four images using the following four random codes:
\begin{align} \label{eq:codes}
&z_{11} = (z_{c1}, z_{s1}), \qquad
z_{12} = (z_{c1}, z_{s2}), \\ \nonumber
&z_{21} = (z_{c2}, z_{s1}), \qquad
z_{22} = (z_{c2}, z_{s2}).
\end{align}
Since the content code is shared between $z_{11}$ and $z_{12}$, the network needs to synthesize two images with the same content for these codes. Nevertheless, since the style codes are different, there should be a force on the network to render different styles on a unique content. The same discussion is also valid for $z_{21}$ and $z_{22}$. Similarly, for $z_{11}$ and $z_{21}$ (or $z_{12}$ and $z_{22}$), the rendered style on the generated images should be consistent while their contents differ significantly from each other. To this end, motivated by \cite{ulyanov2016texture}, we define our content and style consistency losses.

\textbf{Content Consistency Loss:} To learn complex cross-domain relationships, we impose the Euclidean distance on the high-level feature space as the cycle consistency, which is known as the \textit{perceptual loss} \cite{johnson2016perceptual}. Similar to \cite{johnson2016perceptual}, we make use of a VGG-19 network, $\Phi$, pretrained, as a fixed loss network. This loss network defines a feature reconstruction loss that measure differences in high-level content between images. Let $\Phi_l(x)$ denote the feature maps of the $l^{th}$ layer of the loss network for the input image $x$. Then the $l^{th}$ layer content loss between images $x$ and $y$ is defined as:
\begin{align} \label{eq:percept}
L_{c}(x,y) = \frac{1}{N_l}\parallel \Phi_l(x) - \Phi_l(y) \parallel_2^2,
\end{align}
where $N_l=C_l×H_l×W_l$ is the number of perceptrons in the $l^{th}$ layer. Consequently, to train the generator keeping the content between $z_{ii}$ and $z_{ij}$ we define the content-consistency loss as: 
\begin{align} \label{eq:cc_loss}
L_{cc}(z_{ii}, z_{ij}) = L_{c}(G(z_{ii}),G(z_{ij})).
\end{align}

\textbf{Style Consistency Loss:} We also wish to penalize differences in style: colors, textures, common patterns, etc. To achieve this effect, Gatys et al. \cite{gatys2016image} proposed a style reconstruction loss.
As above, let $\Phi_l(x)$ be the activations at the $l^{th}$ layer of the loss network $ϕ$ for the input $x$, which is a feature map of size $C_l\times H_l\times W_l$. By reshaping $\Phi_l(x)$ into a matrix $\varPsi$ of size $C_l\times H_lW_l$, the Gram matrix $GM^l(x)$, which is of size $C_l\times C_l$, can be computed efficiently as $GM^l = \varPsi \varPsi^T/N_l$. The style reconstruction loss is then the squared Frobenius norm of the difference between the Gram matrices of the output and target images:
\begin{align} \label{eq:style}
L_{s}^l(x,y) = \parallel GM^l(x) - GM^l(y) \parallel_F^2.
\end{align}
Finally, to train the generator to share the same style between $z_{ii}$ and $z_{ji}$ we define the style-consistency loss as: 
\begin{align} \label{eq:sc_loss}
L_{sc}(z_{ii}, z_{ji}) = \sum_{l \in L} L_{s}^l(G(z_{ii}),G(z_{ji})),
\end{align}
where $L$ is the set of layers over which  the style-consistency loss is computed. 

\textbf{Diversity Loss:}
Early experiments showed that employing only style and content consistency losses results in the network to learn a limited number of styles. To sidestep this issue, we propose to force a minimum distance in style or content metrics if two images do not share the same style or content code, respectively. In other words, since the style codes differ between $z_{11}$ and $z_{12}$, the network needs to synthesize two images with at least a minimum style difference. To this end, we define the following loss function:
\begin{align} \label{eq:s_div}
L_{sd}(z_{ii}, z_{ij}) = \max \Big(0, m_s-L_{sc}(z_{ii}, z_{ij}) \Big),
\end{align}
where $m_s$ is a margin which is greater than 0 and indicates that dissimilar pairs that are beyond the margin will not contribute to the loss. A similar loss function can be defined for codes with dissimilar content parts and a shared style:
\begin{align} \label{eq:c_div}
L_{cd}(z_{ii}, z_{ji}) = \lambda_{cd} \max \Big(0, m_c-L_{cc}(z_{ii}, z_{ji}) \Big).
\end{align}

\textbf{Total Loss:}
We solve the problem of style and content disentanglement by training the generator $G$ to minimize the following objective function:
\begin{align} \label{eq:total}
L_t &= \lambda_{cc} \big ( L_{cc}(z_{11}, z_{12}) + L_{cc}(z_{21}, z_{22}) \big ) \\ \nonumber & + \lambda_{sc} \big ( L_{sc}(z_{11}, z_{21}) + L_{sc}(z_{12}, z_{22}) \big ) \\ \nonumber & + \lambda_{sd} \big ( L_{sd}(z_{11}, z_{12}) + L_{sd}(z_{21}, z_{22}) \big ) \\ \nonumber & + \lambda_{cd} \big ( L_{cd}(z_{11}, z_{21}) + L_{cd}(z_{12}, z_{22}) \big )\\ \nonumber &+ \lambda_d \sum_{i,j \in \{1, 2\}} L_d(G(z_{ij})),
\end{align}
where the hyper-parameter $\lambda_d, \lambda_{cc}, \lambda_{sc}, \lambda_{sd}, \lambda_{cd}$ controls the impact of each term in the objective function. This objective function forces the generator to disentangle its input code into content and style codes. Note that all the blocks of the generator are trained jointly using the proposed objective function. The discriminator is then trained to maximize the GAN terms $\sum_{i,j \in \{1, 2\}} L_d(G(z_{ij}))$.

\section{Experiments}
In this paper, we employ a Least Squares Generative Adversarial Network (LSGAN) \cite{mao2017least} as the baseline. The LSGAN generator includes six strided convolutional layers. All the layers double the spatial size of their inputs except the first layer which quadruples the input, that leaves us with an image size of $128 \times 128$. We modify the LSGAN to create SC-GAN by adding a stylization block before its last strided convolutional layer. The stylization block consists of 4 residual blocks each equipped with an AdaIN normalization layer. We also replace the normalization layers of the LSGAN as described in section \ref{sec:network}. The style code MLP comprises 5 fully-connected layers with ReLu activation function.

\subsection{Experimental setup}\label{sec:setup}
The main goal of our experiments is to investigate if the proposed method can learn disentangled style and code representations of the data. The framework should be able to learn a content code which controls the geometrical information of the generated image and has no effect on the rendered styles. Similarly, a style representation should be learned to control only the style of the synthesized images. We evaluate SC-GAN on the edges $\leftrightarrow$ handbags \cite{zhu2016generative}, edges $\leftrightarrow$ shoes \cite{yu2014fine}, CelebA \cite{liu2015faceattributes}, LSUN bedroom and kitchen \cite{radford2015unsupervised}, and our own face datasets. Our face dataset has 3,000 face images of high quality. We conducted a comprehensive set of experiments, to evaluate the proposed SC-GAN qualitatively and quantitatively.

\textbf{Disentangled style and content image generation employing LSGAN \cite{mao2017least}}: 
We employ LSGAN in its unconditional setting to conduct the first series of our experiments, as it can efficiently generate realistic and diverse images in different domains. Furthermore, our proposed framework is applicable to any other GAN models. 
In this work, we only modify the generator of LSGAN by adding a stylization block, consisting of four residual blocks, before its last constitutional layer. As we mentioned before, all the layers of this generator, up to the stylization block, are equipped with the IN layer. 
\begin{figure}[t]
	\begin{center}
		
		\includegraphics[width=0.99\linewidth]{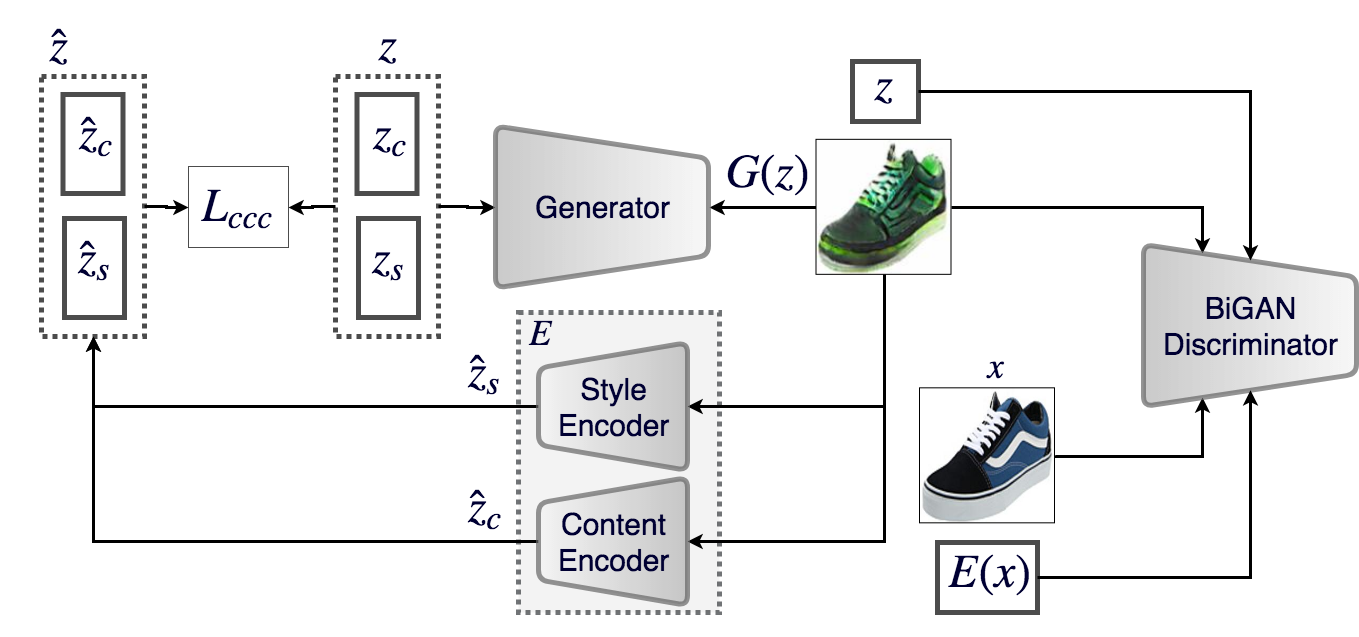}
	\end{center}
	\caption{The proposed SC-GAN framework for style transfer incorporating BiGAN.}
	\label{fig:bigan}
\end{figure}
 
\textbf{Style transfer using BiGAN \cite{donahue2016adversarial} and LSGAN}: Bidirectional Generative Adversarial Network (BiGAN) is an unsupervised feature learning framework. It makes use of an encoder $E$, in addition to the generator $G$, which maps the data $x$ to its latent representations $z$. Its discriminator also differs from the original GAN framework in that the BiGAN discriminator does not discriminate
in data space ($x$ versus $G(z)$). In contrast, it discriminates jointly in the data and latent space (tuples $(x, E(x))$ versus $(G(z), z))$. In other words, the BiGAN encoder $E$ learns to invert the generator $G$. Our early experiments show that this framework, at least for the task of style code retrieval, can benefit from a code cycle-consistency loss defined as:
\begin{align} \label{eq:ccc}
L_{ccc} = \parallel z - E(G(z)) \parallel_2^2.
\end{align}

In order to generate high-quality images, we incorporate the BiGAN framework into the generator of LSGAN. The learned encoder then can be used to retrieve the style code of a given image. The extracted style code may be used, together with a content code, to generate an image with an arbitrary content and the style of a reference image. We used two separate encoders $E_c$ and $E_s$ for content and style retrieval, respectively (see Figure~\ref{fig:bigan}). The content encoder downsamples the input image using several strided convolutional layers, each followed by an IN. Style encoder follows the same structure, however, does not equipped with the IN as it removes the style information.

\begin{figure*}[t!]
	\centering
		\begin{subfigure}[t]{0.33\textwidth}
		\centering
	\end{subfigure}%
	\hfill
	\begin{subfigure}[t]{0.33\textwidth}
		\centering
		\includegraphics[width=.95\linewidth]{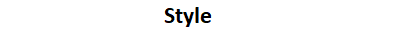}
	\end{subfigure}%
	\hfill
	\begin{subfigure}[t]{0.33\textwidth}
		\centering
	\end{subfigure}

	\begin{subfigure}[t]{0.02\textwidth}
	\centering
	\includegraphics[width=.50\linewidth]{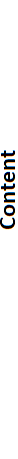}
	\end{subfigure}%
	\begin{subfigure}[t]{0.32\textwidth}
		\centering
		\includegraphics[width=.95\linewidth]{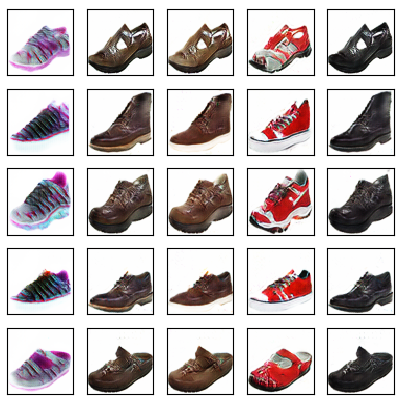}
		\caption{Edges$\leftrightarrow$Shoes.}
		\label{fig:shoes}
	\end{subfigure}%
	\hfill
	\begin{subfigure}[t]{0.32\textwidth}
		\centering
		\includegraphics[width=.95\linewidth]{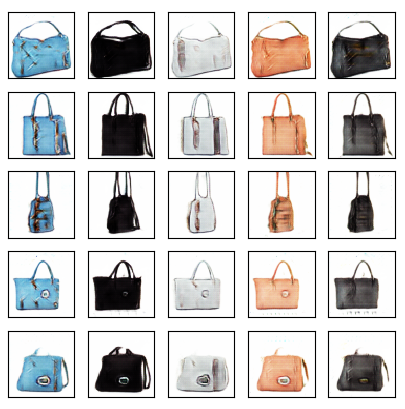}
		\caption{Edges$\leftrightarrow$Handbags}
		\label{fig:bags}
	\end{subfigure}%
	\hfill
	\begin{subfigure}[t]{0.32\textwidth}
		\centering
		\includegraphics[width=.95\linewidth]{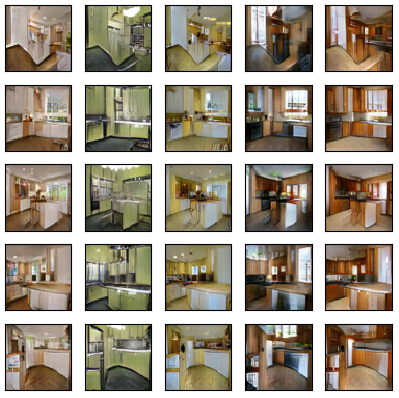}
		\caption{LSUN-kitchen}
		\label{fig:kitchen}
	\end{subfigure}
	
	\begin{subfigure}[t]{0.02\textwidth}
		\centering
		\includegraphics[width=.50\linewidth]{content.png}
	\end{subfigure}%
	\begin{subfigure}[t]{0.32\textwidth}
		\centering
		\includegraphics[width=.95\linewidth]{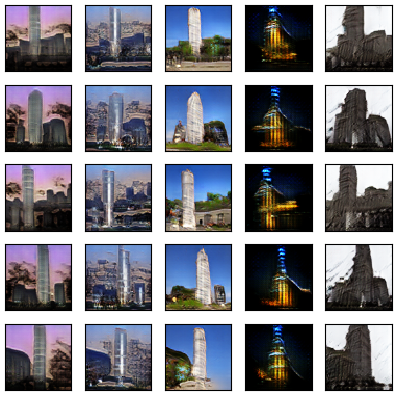}
		\caption{LSUN-tower}
		\label{fig:celeba}
	\end{subfigure}%
	\hfill
	\begin{subfigure}[t]{0.32\textwidth}
		\centering
		\includegraphics[width=.95\linewidth]{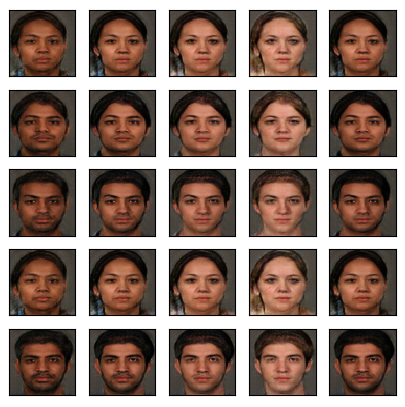}
		\caption{Our face dataset}
		\label{fig:oranges}
	\end{subfigure}
	\hfill
	\begin{subfigure}[t]{0.32\textwidth}
		\centering
		\includegraphics[width=.95\linewidth]{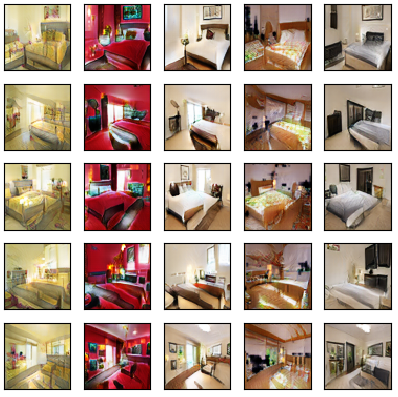}
		\caption{LSUN-bedroom}
		\label{fig:reverse}
	\end{subfigure}
	\caption{The results of our framework on different datasets. The content code is fixed in each row while the style code varies. Similarly, the style code is fixed in each column while the content code varies.}
	\label{fig:results}
\end{figure*}
We train our proposed networks using Adam optimizer \cite{kingma2014adam}, with learning rate of 0.0002, $\beta_1=0.5$, $\beta_2=0.999$ and mini-batch size of 20. The algorithm is implemented in PyTorch \cite{paszke2017automatic}. The LSGAN is trained to generate images of size $128\times 128$. The SC-GAN is quite robust to the values of hyper-parameters. Selecting $\lambda_d = 0.5$, $\lambda_{cc} = 1$, $\lambda_{sc}= 4$, $\lambda_{sd} = 1$, $\lambda_{cd}=1$, $m_s=1$ and $m_c=0.5$ easily works for all the datasets.

\subsection{Qualitative Analysis} \label{sec:qualitative}

Figure~\ref{fig:results} show how the generated images change when the style or content code varies. The content code is fixed in each row while the style code differs between different columns. Similarly, we fixed the style code in each column. Our framework is able to train a network to generate multiple realistic images with fixed content and different styles, while it does not require any supervision. Figure~\ref{fig:results} clearly illustrates the disentanglement of the content and style representations for different datasets.

\begin{figure*}[t!]
	\centering
	\hspace*{\fill}
	\begin{subfigure}[t]{0.40\textwidth}
		\centering
		\includegraphics[width=.95\linewidth]{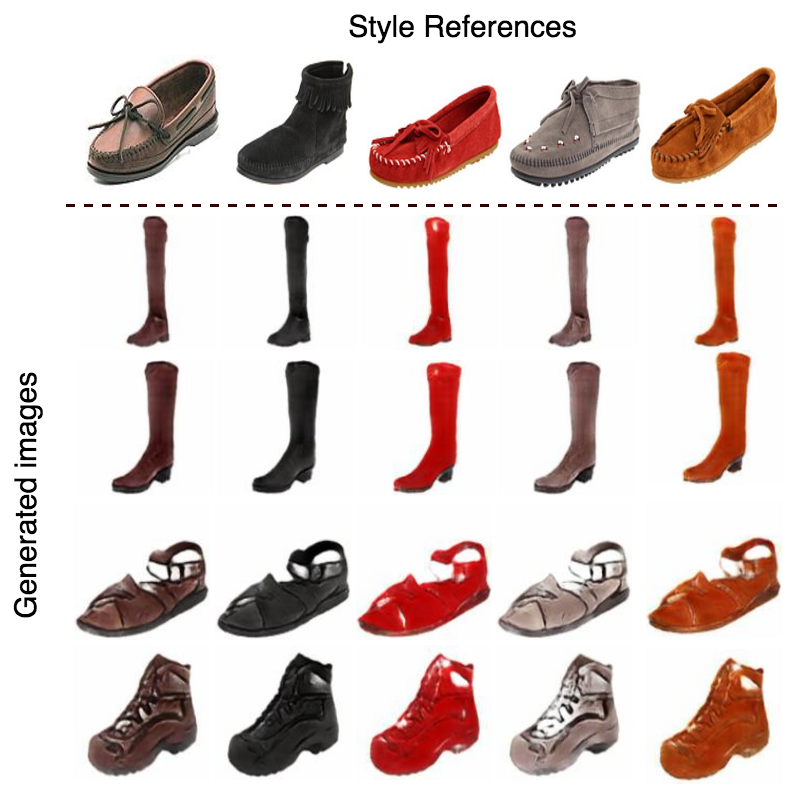}
		\caption{Edges$\leftrightarrow$Shoes.}
		\label{fig:shoes2}
	\end{subfigure}%
	\hfill
	\begin{subfigure}[t]{0.40\textwidth}
		\centering
		\includegraphics[width=.95\linewidth]{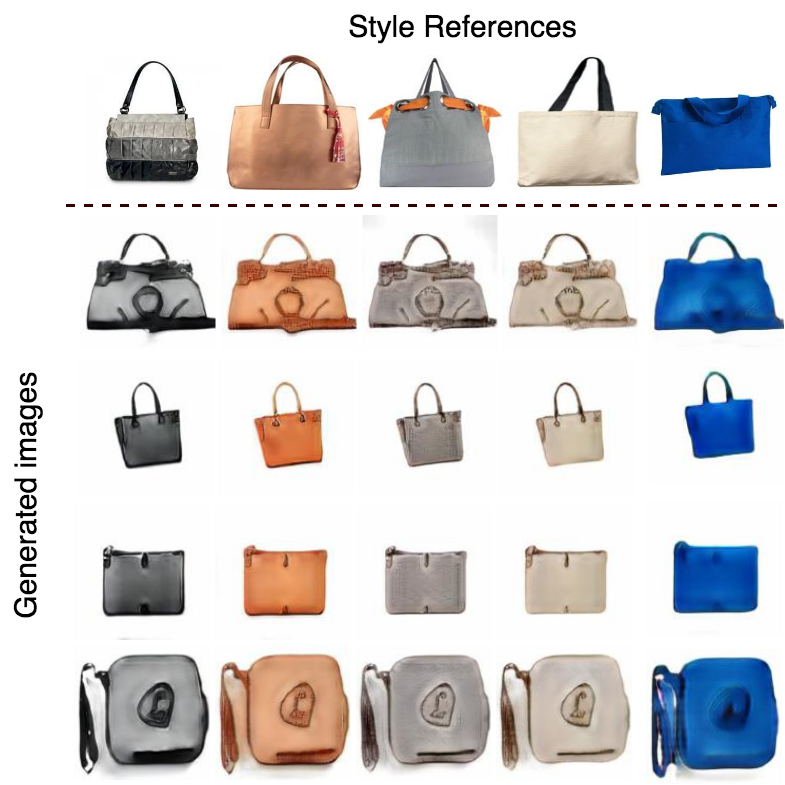}
		\caption{Edges$\leftrightarrow$Handbags}
		\label{fig:bags2}
	\end{subfigure}
	\hspace*{\fill}
	\caption{Training SC-GAN with an encoder to retrieve the style code enables us to transfer style from a given reference image to the generated samples.}
	\label{fig:transfer}
\end{figure*}

Furthermore, using the proposed style transfer scheme, we can transfer style information from a reference image to the generated image by a random content code. To this end, we jointly train an encoder with the generator to retrieve the style code from an image along with the training domain. Then, instead of sampling from the distribution of style code, we use the style code extracted from a reference image. To this end, the reference image is fed to the learned encoder, $E$, to retrieve its style code. The extracted code then can be utilized for image generation guided by the reference image style. Figure~\ref{fig:transfer} shows the results using style codes extracted from multiple reference images to generate realistic photos in different domains. 

\subsection{Quantitative Evaluation}
\begin{table*}[]
	\centering
	\begin{tabular}{|c|c|c|c|c|c|c|c|}
		\hline
		\multicolumn{2}{|c|}{Dataset}   & Handbags & Shoes & Bedroom & Kitchen & Tower & Face \\ \hline
		\multirow{3}{*}{FID}   & LSGAN  &   84.13     &   69.28    &   75.02      &    80.62     &   69.34     &  52.83    \\ \cline{2-8} 
		& SC-GAN w/o Diversity Loss &    81.13      &  65.49     &   74.08      &    77.80     &   68.14     &  50.99    \\ \cline{2-8}
		& SC-GAN w/ Diversity Loss &    80.51      &  65.13     &   71.97      &    78.53     &   70.90     &  50.29    \\ \hline
		\multirow{3}{*}{LPIPS} & LSGAN  &  0.101       &   0.124    &   0.119      &     0.091    &    0.088    &   0.040   \\ \cline{2-8} 
		& SC-GAN w/o Diversity Loss &    0.109       &   0.119    &  0.127       &    0.095     &   0.090     &  0.038    \\ \cline{2-8}
		& SC-GAN w/ Diversity Loss &    0.132       &   0.164    &  0.143       &    0.109     &   0.094     &  0.051    \\ \hline
	\end{tabular}
	\caption{Comparison of LSGAN and SC-GAN, with and without Diversity Loss, on different datasets using the FID and LPIPS scores as quantitative metrics. Lower FID score means higher quality, and higher LPIPS shows more diversity among generated samples.}
	\label{tab:compare}
\end{table*}
To evaluate the proposed method quantitatively, we use two different metrics, namely Frechet Inception Distance (FID) \cite{heusel2017gans} and Perceptual Image Patch Similarity (LPIPS) \cite{zhang2018unreasonable}. We trained the original LSGAN and our SC-GAN extension of that on multiple datasets and then compared their performances based on FID and LPIPS metrics.

The FID is a recently proposed metric to evaluate the quality of the generative models \cite{heusel2017gans}. It directly measures the distance between the synthetic data distribution $p(.)$ and the real data distribution $p_r(.)$. To calculate FID, images are encoded with visual features from a pre-trained inception model: 
\begin{align} \label{eq:fid}
FID = \parallel m-m_r \parallel_2^2 Tr(C+C_r-2(CC_r)^{1/2}) ,
\end{align}
where $(m, C)$ and $(m_r, C_r)$ denote the mean and covariance of feature embedding for synthetic and real data, respectively. Note that a lower FID value interprets as a lower distance between synthetic and real data distributions. We calculate the FID over 10k randomly generated samples. Table~\ref{tab:compare} lists the FID scores of LSGAN and our proposed SC-GAN method. To investigate the effect of diversity loss we train our SC-GAN model with and without diversity loss. Compared to LSGAN, our SC-GAN achieves a slightly better FID. The SC-GAN without diversity loss, however, shows almost the same FID score as LSGAN. It means that the diversity loss can improve the FID score by increasing the diversity of generated sample styles. These results indicate  that the proposed disentanglement of style and content representations comes at least with no cost in terms of the quality of synthesized images. 

\begin{figure}[t!]
	\centering
	\begin{subfigure}[t]{0.40\textwidth}
		\centering
		\includegraphics[width=.95\linewidth]{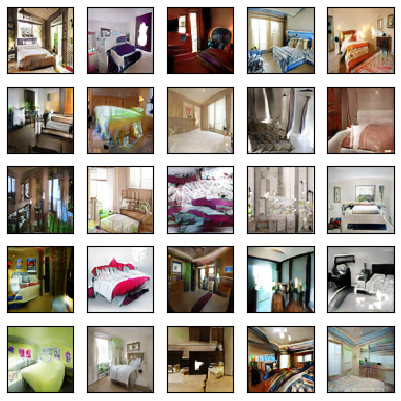}
		\caption{With Diversity Loss.}
		\label{fig:with}
	\end{subfigure}%
	\hfill
	\begin{subfigure}[t]{0.40\textwidth}
		\centering
		\includegraphics[width=.95\linewidth]{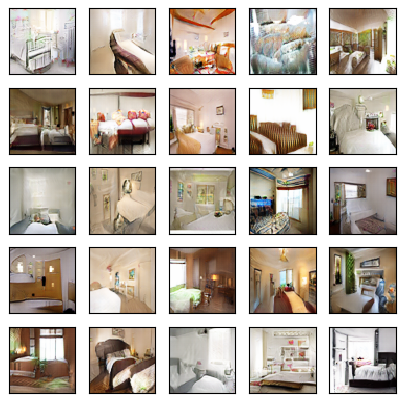}
		\caption{Without Diversity Loss.}
		\label{fig:without}
	\end{subfigure}
	\caption{Results of a SC-GAN trained on LSUN-bedroom dataset with and without diversity loss. Employing the loss improves the style diversity.}
	\label{fig:with_without}
\end{figure}

The observations regarding the diversity of generated samples are confirmed by the LPIPS distance. Table~\ref{tab:compare} presents the LPIPS distance for LSGAN as well as SC-GAN with and without diversity loss. The LPIPS distance is calculated as the average distance between 2000 pairs of randomly generated output images, in deep feature space of a pre-trained AlexNet \cite{krizhevsky2012imagenet}. Without the diversity loss, our model suffers from partial mode collapse, in which many style codes render the same texture on the output images. However, employing the diversity loss results in the SC-GAN to generate images that are more diverse. Visualization comparison of SC-GAN with and without diversity loss in Figure~\ref{fig:with_without} confirms the LPIPS scores. Note that the proposed SC-GAN has a higher LPIPS score compared to the LSGAN while it is still able to generate images of almost the same quality. 

We have also investigated different generator architectures to find out where is the best place for the stylization block. The studied architectures differ only in how early we use the stylization block. We tried three different architectures: c5-r4-c1, c4-r4-c2, and c3-r4-c3. Here, c$i$-r$j$-c$k$ denote a generator with decoder \#1, residual blocks, and decoder \#2 having $i$, $j$, and $k$ layers, respectively. Table~\ref{tab:diff_structure} shows the FID for these architectures on three different datasets. Clearly, using stylization block in the first layers reduces the performance of the SC-GAN. We achieved the best performance by placing the stylization block before the last convolutional layer (c5-r4-c1). Note that all the results in Table~\ref{tab:compare} are reported for this architecture.

\begin{table}[]
	\begin{tabular}{|c|c|c|c|c|}
		\hline
		& Handbags & Shoes & Bedroom & Kitchen \\ \hline
		c5-r4-c1 &   80.51 & 65.13   &   71.97    &   78.53         \\ \hline
		c4-r4-c2 &   85.64 & 68.43   &   76.12    &   80.33 \\ \hline
		c3-r4-c3 &   89.13 & 70.20   &   81.97    &   87.51 \\ \hline
	\end{tabular}
	\caption{FID score for different generator architectures. The structures differ only in how early we use the stylization block. Clearly, using stylization block in the early layers reduce the performance of the GAN.}
	\label{tab:diff_structure}
\end{table}

\section{Conclusion}
In this paper, we introduced a framework for style and content disentangled representation learning, namely SC-GAN, using generative adversarial networks in an unsupervised setting. In contrast to the previous works, our approach learns distinct content and style codes for a given dataset, which enables us to generate multiple images of a scene with different styles and textures. We also proposed to retrieve the style code which may be used later for style transfer from a given reference image to the generated image. The proposed SC-GAN can easily be adopted by any GAN frameworks. Extensive quantitative and qualitative results demonstrate that our proposed method can learn to disentangle representations of style and content while improving the quality of the generated images.

{\small
\bibliographystyle{ieee}
\bibliography{egbib}
}

\end{document}